\documentclass[10pt,twocolumn,letterpaper]{article}

\usepackage{iccv}
\usepackage{graphicx}
\usepackage{times}
\usepackage{epsfig}
\usepackage{amsmath}
\usepackage{amssymb}
\usepackage{algorithm}
\usepackage{multirow}
\usepackage{mathtools,xparse}
\usepackage{siunitx}
\usepackage{stmaryrd}
\usepackage{amsmath}
\usepackage[noend]{algpseudocode}
\usepackage{placeins}
\usepackage{mdframed}
\usepackage{xspace}
\usepackage{amsmath,amssymb} 
\usepackage{color}
\usepackage{subcaption}
\usepackage[table]{xcolor}

\newcommand{\trans}{\text{T}}

\definecolor{g1}{rgb}{1.00,1.00,1.00}

\newcommand{\matr}[1]{\mathbf{#1}}     

\newcommand{\Hom}{\matr{H}}
\newcommand{\Aff}{\matr{A}}

\newcommand{\Point}{\matr{p}}

\def\gb{Gr{\"o}bner basis\xspace}

\newcommand{\ph}[1]{\phantom{#1}}
\definecolor{Gray}{gray}{0.85}
\newcolumntype{g}{>{\columncolor{Gray}}c}

\usepackage[pagebackref=true,breaklinks=true,letterpaper=true,colorlinks,bookmarks=false]{hyperref}

\iccvfinalcopy 
 
\def\iccvPaperID{2962} 

\begin{document}

\title{Homography from two orientation- and scale-covariant features}

\author{Daniel Barath$^{12}$, and Zuzana Kukelova$^{1}$\\
$^1$ Centre for Machine Perception, Department of Cybernetics \\
  Czech Technical University, Prague, Czech Republic \\
  $^2$ Machine Perception Research Laboratory, 
  MTA SZTAKI, Budapest, Hungary \\
    {\tt\small barath.daniel@sztaki.mta.hu}
}


\maketitle

\begin{abstract}
This paper proposes a geometric interpretation of the angles and scales which the orientation- and scale-covariant feature detectors, e.g.\ SIFT, provide. 
Two new general constraints are derived on the scales and rotations which can be used in any geometric model estimation tasks.
Using these formulas, two new constraints on homography estimation are introduced. 
Exploiting the derived equations, a solver for estimating the homography from the minimal number of two correspondences is proposed.
Also, it is shown how the normalization of the point correspondences affects the rotation and scale parameters, thus achieving numerically stable results.
Due to requiring merely two feature pairs, robust estimators, e.g.\ RANSAC, do significantly fewer iterations than by using the four-point algorithm.
When using covariant features, e.g.\ SIFT, the information about the scale and orientation is given at no cost. 
The proposed homography estimation method is tested in a synthetic environment and on publicly available real-world datasets.    
\end{abstract}

\section{Introduction}

This paper addresses the problem of interpreting, in a geometrically justifiable manner, the rotation and scale which the orientation- and scale-covariant feature detectors, e.g.\ SIFT \cite{lowe1999object} or SURF \cite{bay2006surf}, provide. Then, by exploiting these new constraints, we involve all the obtained parameters of the SIFT features (i.e.\ the  point coordinates, angle, and scale) into the homography estimation procedure. 
In particular, we are interested in the minimal case, to estimate a homography from solely two correspondences.   

Nowadays, a number of algorithms exist for estimating or approximating geometric models, e.g.\ homographies, using affine-covariant features. 
A technique, proposed by Perdoch et al.~\cite{PerdochMC06}, approximates the epipolar geometry from one or two affine correspondences by converting them to point pairs.   
Bentolila and Francos~\cite{Bentolila2014} proposed a solution for estimating the fundamental matrix using three affine features. 
Raposo et al.~\cite{Raposo2016} and Barath et al.~\cite{barath2018efficient} showed that two correspondences are enough for estimating the relative camera motion. 
Moreover, two feature pairs are enough for solving the semi-calibrated case, i.e.\ when the objective is to find the essential matrix and a common unknown focal length~\cite{barath2017focal}. 
Also, homographies can be estimated from two affine correspondences~\cite{koser2009geometric}, and, in case of known epipolar geometry, from a single correspondence~\cite{barath2017theory}. 
There is a one-to-one relationship between local affine transformations and surface normals~\cite{koser2009geometric, barath2015optimal}. 
Pritts et al.~\cite{Pritts2017RadiallyDistortedCT} showed that the lens distortion parameters can be retrieved using affine features. 
\begin{figure}[t]
	\centering
    \includegraphics[width = 0.80\columnwidth]{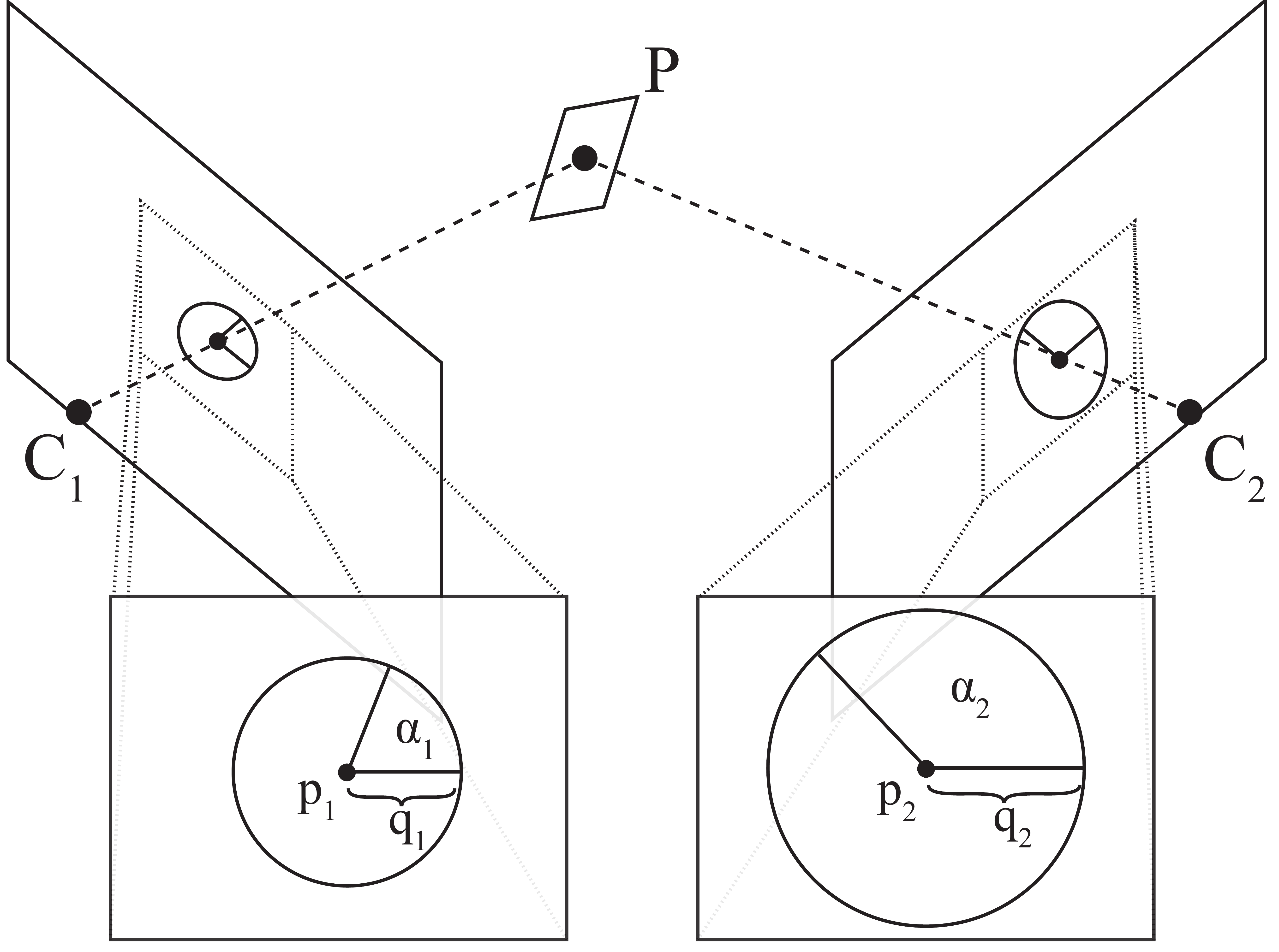}
    \caption{ Visualization of the orientation- and scale-covariant features. Point $\textbf{P}$ and the surrounding patch projected into cameras $\textbf{C}_1$ and $\textbf{C}_2$. A window showing the projected points $\textbf{p}_1 = [u_1 \; v_1 \; 1]^\trans$ and $\textbf{p}_2 = [u_2 \; v_2 \; 1]^\trans$ are cut out and enlarged. The rotation of the feature in the $i$th image is $\alpha_i$ and the size is $q_i$ ($i \in \{1,2\}$). The scaling from the $1$st to the $2$nd image is calculated as $q = q_2 / q_1$. }
    \label{fig:geometric_interpr_sift}
\end{figure}
Affine correspondences encode higher-order information about the scene geometry. This is the reason why the previously mentioned algorithms solve geometric estimation problems exploiting fewer features than point correspondence-based methods. This implies nevertheless their major drawback: obtaining affine features accurately (e.g.\ by Affine SIFT \cite{morel2009asift}, MODS \cite{mishkin2015mods}, Hessian-Affine, or Harris-Affine \cite{mikolajczyk2005comparison} detectors) is time-consuming and, thus, is barely doable in time-sensitive applications.

Most of the widely-used feature detectors provide parts of the affine feature. 
For instance, there are detectors obtaining oriented features, e.g.\ ORB~\cite{rublee2011orb}, or there are ones providing also the scales, e.g.\ SIFT~\cite{lowe1999object} or SURF~\cite{bay2006surf}.  
Exploiting this additional information is a well-known approach in, for example, wide-baseline matching~\cite{matas2004robust,mishkin2015mods}. Yet, the first papers~\cite{barath2017phaf,barath2018approximate,barath2018five,mills2018four,barath2018recovering} involving them into geometric model estimation were published just in the last few years. 
In \cite{mills2018four}, the feature orientations are involved directly in the essential matrix estimation. 
In \cite{barath2017phaf}, the fundamental matrix is assumed to be a priori known and an algorithm is proposed for approximating a homography exploiting the rotations and scales of two SIFT correspondences. The approximative nature comes from the assumption that the scales along the axes are equal to the SIFT scale and the shear is zero. In general, these assumptions do not hold. 
The method of \cite{barath2018approximate} approximates the fundamental matrix by enforcing the geometric constraints of affine correspondences on the epipolar lines. Nevertheless, due to using the same affine model as in \cite{barath2017phaf}, the estimated epipolar geometry is solely an approximation. 
In \cite{barath2018five}, a two-step procedure is proposed for estimating the epipolar geometry. First, a homography is obtained from three oriented features. Finally, the fundamental matrix is retrieved from the homography and two additional correspondences. Even though this technique considers the scales and shear as unknowns, thus estimating the epipolar geometry instead of approximating it, the proposed decomposition of the affine matrix is not justified theoretically. Therefore, the geometric interpretation of the feature rotations is not provably valid. 
A recently published paper~\cite{barath2018recovering} proposes a way of recovering full affine correspondences from the feature rotation, scale, and the fundamental matrix. Applying this method, a homography is estimated from a single correspondence in case of known epipolar geometry. Still, the decomposition of the affine matrix is ad hoc, and is, therefore, not a provably valid interpretation of the SIFT rotations and scales. Moreover, in practice, the assumption of the known epipolar geometry restricts the applicability of the method. 

The contributions of this paper are: (i) we provide a geometrically valid way of interpreting orientation- and scale-covariant features approaching the problem by differential geometry. (ii) Building on the derived formulas, we propose two general constraints which hold for covariant features. (iii) These constraints are then used to derive two new formulas for homography estimation and (iv), based on these equations, a solver is proposed for estimating a homography matrix from two orientation- and scale-covariant feature correspondences. 
This additional information, i.e.\ the scale and rotation, is given at no cost when using most of the widely-used feature detectors, e.g.\ SIFT or SURF.
It is validated both in a synthetic environment and on more than $10\;000$ publicly available real image pairs that the solver accurately recovers the homography matrix. Benefiting from the number of correspondences required, robust estimation, e.g.\ by GC-RANSAC~\cite{barath2017graph}, is two orders of magnitude faster than by combining it with the standard techniques, e.g.\ four-point algorithm~\cite{hartley2003multiple}.     

\section{Theoretical background}

\noindent \textbf{Affine correspondence}
$(\Point_1, \Point_2, \Aff)$ is a triplet, where $\Point_1 = [u_1 \; v_1 \; 1]^\trans$ and $\Point_2 = [u_2 \; v_2 \; 1]^\trans$ are a corresponding homogeneous point pair in two images and $\Aff$
is a $2 \times 2$ linear transformation which is called \textit{local affine transformation}. Its elements in a row-major order are: $a_1$, $a_2$, $a_3$, and $a_4$. To define $\Aff$, we use the definition provided in~\cite{Molnar2014} as it is given as the first-order Taylor-approximation of the $\text{3D} \to \text{2D}$ projection functions. For perspective cameras, the formula for $\Aff$ is the first-order approximation of the related \textit{homography} matrix
as follows: 
\begin{eqnarray}
		\begin{array}{lllllll}
          a_{1} & = & \frac{\partial u_2}{\partial u_1} = \frac{h_{1} - h_{7} u_2}{s}, & & 
          a_{2} & = & \frac{\partial u_2}{\partial v_1} = \frac{h_{2} - h_{8} u_2}{s}, \\[2mm]
          a_{3} & = & \frac{\partial v_2}{\partial u_1} = \frac{h_{4} - h_{7} v_2}{s}, & &
          a_{4} & = & \frac{\partial v_2}{\partial v_1} = \frac{h_{5} - h_{8} v_2}{s}, 
		\end{array}
        \label{eq:taylor_approximation}
\end{eqnarray}
where $u_i$ and $v_i$ are the directions in the $i$th image ($i \in \{1,2\}$) and $s = u_1 h_7 + v_1 h_8 + h_9$ is the projective depth. The elements of $\Hom$ in a row-major order are: $h_1$, $h_2$, ..., $h_9$.

\noindent \textbf{The relationship} of an affine correspondence and a homography is described by six linear equations. Since an affine correspondence involves a point pair, the well-known equations (from $\Hom \Point_1 \sim \Point_2$) hold~\cite{hartley2003multiple}. They are as follows:
\begin{equation}
    \small
    \begin{aligned}
    \label{eq:orig_dlt}
        u_1 h_{1} + v_1 h_{2} + h_{3} - u_1 u_2 h_{7} - v_1 u_2 h_{8} - u_2 h_{9} &= 0, \\
        u_1 h_{4} + v_1 h_{5} + h_{6} - u_1 v_2 h_{7} - v_1 v_2 h_{8} - v_2 h_{9} &= 0.
    \end{aligned}
\end{equation}
After re-arranging (\ref{eq:taylor_approximation}), four additional linear constraints are obtained from $\Aff$ which are the following.
\begin{equation}
    \small
    \begin{aligned}
    	\label{eq:ha}
    	h_{1} - \left( u_2 + a_{1} u_1  \right) h_{7} - a_{1} v_1 h_{8} - a_{1} h_{9} &= 0, \\
    	h_{2} - \left( u_2 + a_{2} v_1  \right) h_{8} - a_{2} u_1 h_{7} - a_{2} h_{9} &= 0, \\
    	h_{4} - \left( v_2 + a_{3} u_1  \right) h_{7} - a_{3} v_1 h_{8} - a_{3} h_{9} &= 0, \\
    	h_{5} - \left( v_2 + a_{4} v_1  \right) h_{8} - a_{4} u_1 h_{7} - a_{4} h_{9} &= 0. 
    \end{aligned}
\end{equation}
Consequently, an affine correspondence provides six linear equations for the elements of the related homography. 

\section{Affine transformation model}

In this section, the interpretation of the feature scales and rotations are discussed.
Two new constraints that relate the elements of the affine transformation to the feature scale and rotation are derived.
These constraints are general, and they can be used for estimating different geometric models, e.g.\ homographies or fundamental matrices, using orientation- and scale-covariant features.
In this paper, the two constraints are used to derive a solver for homography estimation from two correspondences.
For the sake of simplicity, we use SIFT as an alias for all the orientation- and scale-covariant detectors. The formulas hold for all of them.

\subsection{Interpretation of the SIFT output}

Reflecting the fact that we are given a scale $q_i \in \mathbb{R}$ and rotation $\alpha_i \in [0, 2 \pi)$ independently in each image ($i \in \{ 1, 2 \}$; see Fig.~\ref{fig:geometric_interpr_sift}), the objective is to define affine correspondence $\Aff$ as a function of them. 
For this problem, approaches were proposed in the recent past~\cite{barath2018five,barath2018recovering}. None of them were nevertheless proven to be a valid interpretation.   

To understand the SIFT output, we exploit the definition of affine correspondences proposed in~\cite{barath2015optimal}. In~\cite{barath2015optimal}, $\Aff$ is defined as the multiplication of the Jacobians of the projection functions in the two images as follows:
\begin{equation}
    \textbf{A} = \matr{J}_2 \matr{J}_1^{-1},
    \label{eq:A_as_jacobians}
\end{equation}
where $\matr{J}_1$ and $\matr{J}_2$ are the Jacobians of the 3D $\to$ 2D projection functions. Proof is in Appendix~\ref{app:proof_jacobians}.
For the $i$th Jacobian, the following is a possible decomposition:
\begin{equation}
    \small
    \textbf{J}_i = \textbf{R}_i \textbf{U}_i = \begin{bmatrix} 
        \cos(\alpha_i) & -\sin(\alpha_i) \\
        \sin(\alpha_i) & \cos(\alpha_i)
    \end{bmatrix} 
    \begin{bmatrix} 
        q_{u,i} & w_i \\
        0 & q_{v,i}
    \end{bmatrix},
\end{equation}
where angle $\alpha_i$ is the rotation in the $i$th image, $q_{u,i}$ and $q_{v,i}$ are the scales along axes $u$ and $v$, and $w_i$ is the shear ($i \in \{1,2\}$). Let us use the following notation: $c_i = \cos(\alpha_i)$ and $s_i = \sin(\alpha_i)$.
The equation for the inverse matrix becomes
\begin{eqnarray*}
    \small
    \textbf{J}_i^{-1} = 
    \frac{1}{c_i^2 q_{u,i} q_{v,i} + s_i^2 q_{u,i} q_{v,i}} \begin{bmatrix}
        s_i w_i + c_i q_{v,i} & s_i q_{v,i} - c_i w_i \\
        -s_i q_{u,i} & c_i q_{u,i}
    \end{bmatrix}.
\end{eqnarray*}
The denominator can be formulated as follows: $(c_i^2 + s_i^2) q_{u,i} q_{v,i}$, where $c_i^2 + s_i^2$ is a trigonometric identity and equals to one.
After multiplying the matrices in (\ref{eq:A_as_jacobians}), the following equations are given for the affine elements: 
\begin{eqnarray}
    \small
    \label{eq:constraints_a_qu_qv_1}
    a_1 = \frac{c_2 q_{u,2} (s_1 w_1 + c_1 q_{v,1}) - s_1 q_{u,1} (c_2 w_2 - s_2 q_{v,2})}{q_{u,1} q_{v,1}} \\
    a_2 = \frac{c_2 q_{u,2} (s_1 q_{v,1} - c_1 w_1) + c_1 q_{u,1} (c_2 w_2 - s_2 q_{v,2})}{q_{u,1} q_{v,1}} \\
    a_3 = \frac{s_2 q_{u,2} (s_1 w_1 + c_1 q_{v,1}) - s_1 q_{u,1} (s_2 w_2 + c_2 q_{v,2})}{q_{u,1} q_{v,1}} \\
    a_4 = \frac{s_2 q_{u,2} (s_1 q_{v,1} - c_1 w_1) + c_1 q_{u,1} (s_2 w_2 + c_2 q_{v,2})}{q_{u,1} q_{v,1}} 
    \label{eq:constraints_a_qu_qv_4}
\end{eqnarray}
These formulas show how the affine elements relate to $\alpha_i$, the scales along axes $u$ and $v$ and shears $w_i$.

In case of having orientation- and scale-covariant features, e.g.\ SIFT, the known parameters are the rotation $\alpha_i$ of the feature in the $i$th image and a uniform scale $q_i$. It can be easily seen that the scale $q_i$ is interpreted as follows:
\setlength{\abovedisplayskip}{5pt}
\setlength{\belowdisplayskip}{2pt}
\setlength{\abovedisplayshortskip}{0pt}
\setlength{\belowdisplayshortskip}{0pt}
\begin{equation}
    q_i = \det \matr{J}_i = q_{u,i} q_{v,i}. 
    \label{eq:scale}
\end{equation}
Therefore, our goal is to derive constraints that relate affine elements of $\Aff$ to the orientations $\alpha_i$ and scales $q_i$ of the features in the first and second images.
We will derive such constraints by eliminating the scales along axes $q_{u,i}$ and $q_{v,i}$ and the shears $w_i$ from equations~(\ref{eq:constraints_a_qu_qv_1})-(\ref{eq:constraints_a_qu_qv_4}).
To do this, we use an approach based on the elimination ideal theory~\cite{cox2005}. Elimination ideal theory is a classical algebraic method for eliminating variables from polynomials of several variables.
This method was recently used in~\cite{kukelovaCVPR2017} for eliminating unknowns from equations that are not dependent on input measurements. 
Here, we use the method in a slightly different way. We first create the ideal $I$~\cite{cox2005} generated by polynomials~(\ref{eq:constraints_a_qu_qv_1})-(\ref{eq:constraints_a_qu_qv_4}), polynomial~(\ref{eq:scale}) and trigonometric identities $c_i^2 + s_i^2 = 1$ for $i \in \{1,2\}$. Note that here we consider all elements of these polynomials, including $c_i$ and $s_i$, as unknowns. 
Then we compute generators of the elimination ideal $I_1 = I \cap \mathbb{C}[a_1,a_2,a_3,a_4,q_1,q_2,s_1,c_1,s_2,c_2]$~\cite{cox2005}.
The generators of $I_1$ do not contain $q_{u,i}$, $q_{v,i}$ and $w_i$. 
The elimination ideal $I_1$ is generated by two polynomials:
\begin{eqnarray}
    \small
    \label{eq:constraint1}
    q_1^2 a_2 a_3 - q_1^2 a_1 a_4 + q_1 q_2 = 0,\\ 
    \label{eq:constraint2}
    c_1 s_2 q_1 a_1 + s_1 s_2 q_1 a_2 - c_1 c_2 q_1 a_3 - c_2 s_1 q_1 a_4 = 0.
\end{eqnarray}
Generators~(\ref{eq:constraint1})-(\ref{eq:constraint2}) can be computed using a computer algebra system, e.g.\ {\tt Macaulay2}~\cite{M2}. The input code for {\tt Macaulay2} is in the supplementary material.
%
The new constraints relate the elements of $\Aff$ to the scales and rotations of the features in both images. Note that both these equations can be divided by $q_1 \neq 0$. After this simplification, (\ref{eq:constraint1}) corresponds to $\det \Aff = q_2 / q_1 = q$ and equation~(\ref{eq:constraint2}) relates the rotations of the features to the elements of  $\Aff$.
The two new constraints are general, and they can be used for estimating different geometric models, e.g.\ homographies or fundamental matrices, using orientation- and scale-covariant detectors.
Next, we use~(\ref{eq:constraint1})-(\ref{eq:constraint2}) to derive new constraints on a homography.

\begin{figure*}[h] 
    \centering
	\begin{subfigure}[t]{0.495\columnwidth}
	    \includegraphics[width=1.0\columnwidth]{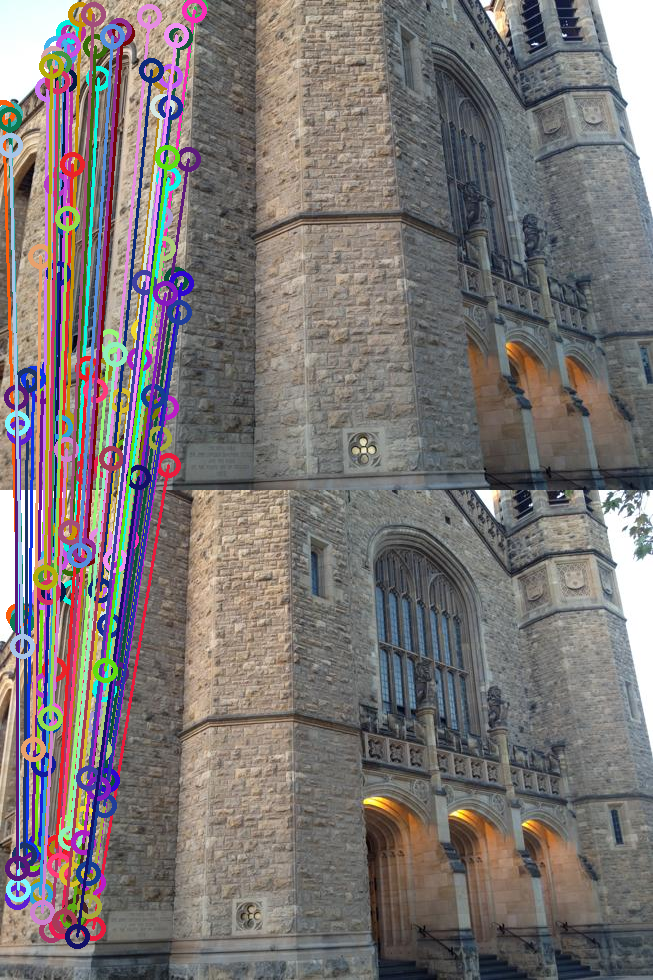}
	    \caption{$553$ iterations by 2SIFT and $8\;615$ by 4PT. Inlier ratio $0.38$. }
	\end{subfigure}\hfill
	\begin{subfigure}[t]{0.495\columnwidth}
	    \includegraphics[width=1.0\columnwidth]{assets/glasscasea_plane_0_matches.png}
	    \caption{$720$ iterations by 2SIFT and $78\;450$ by 4PT. Inlier ratio $0.06$. }
	\end{subfigure}\hfill
	\begin{subfigure}[t]{0.495\columnwidth}
	    \includegraphics[width=1.0\columnwidth]{assets/strecha_castlep19_8_9_matches.png}
	    \caption{$169$ iterations by 2SIFT and $573$ by 4PT. Inlier ratio $0.22$.}
	\end{subfigure}\hfill
	\begin{subfigure}[t]{0.495\columnwidth}
	    \includegraphics[width=1.0\columnwidth]{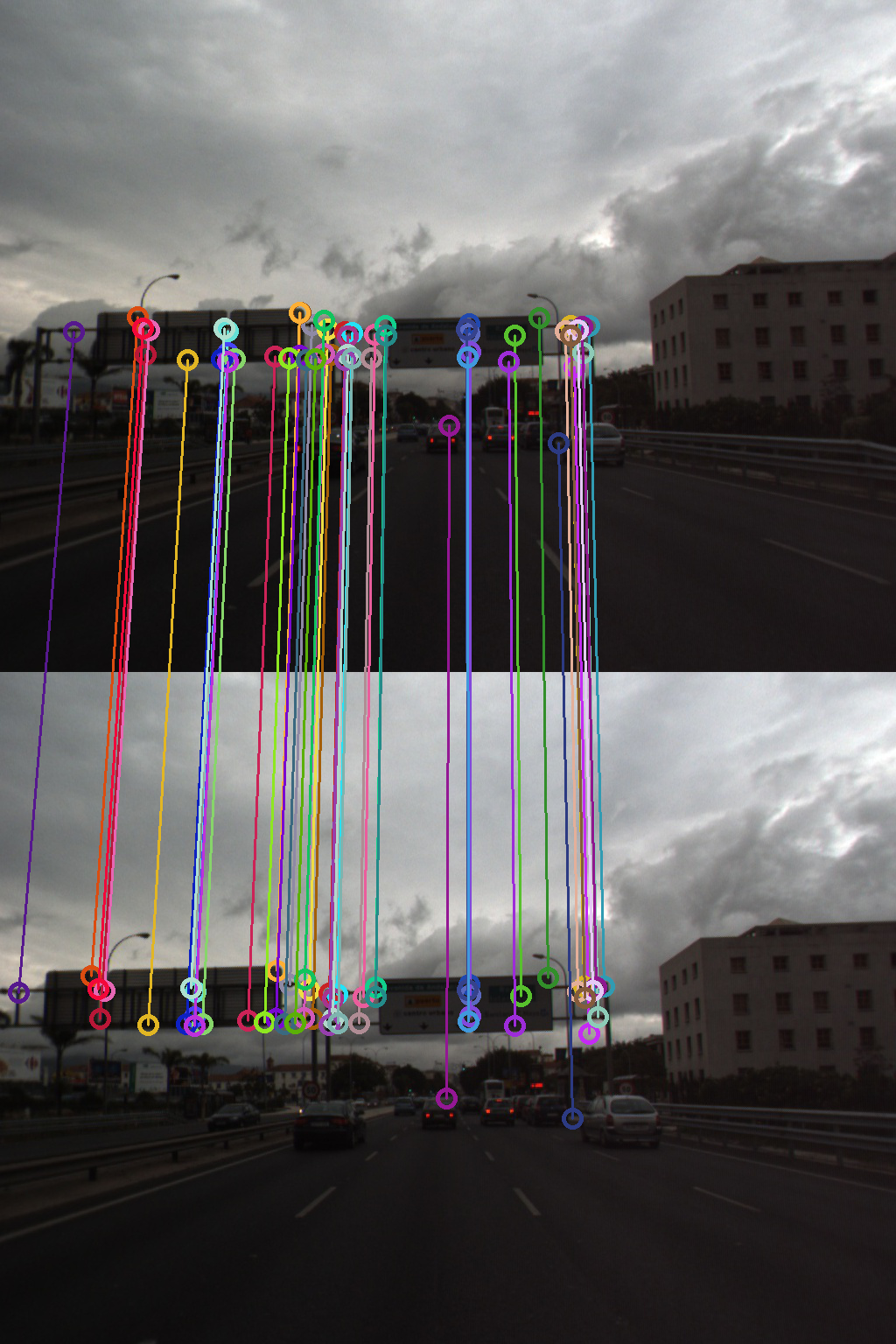}
	    \caption{$65$ iterations by 2SIFT and $14\;139$ by 4PT. Inlier ratio $0.23$.}
	\end{subfigure}
    \caption{ Inliers of the estimated homographies (by 2SIFT) drawn to example image pairs. The numbers of iterations of GC-RANSAC~\cite{barath2017graph} using the 4PT and proposed 2SIFT solvers; and the ground truth inlier ratios are reported in the captions. } 
    \label{fig:example_image_pairs}
\end{figure*} 

\section{Homography from two correspondences}

In this section,  we derive new constraints 
that relates $\Hom$ to the feature scales and rotations in the two images.
Then a solver is proposed to estimate $\Hom$ from two SIFT correspondences based on these new constraints.
Finally, 
we discuss how the widely-used normalization of the point correspondences~\cite{hartley1997defense} affects the output of orientation- and scale-covariant detectors and subsequently the new constraints. 

\subsection{Homography and covariant features}
First, we derive constraints 
that relates the homography $\Hom$ to the scales and rotations of the features in the first and second images.
To do this, we combine constraints~(\ref{eq:constraint1}) and~(\ref{eq:constraint2}) derived in previous section with the constraints on the homography matrix~(\ref{eq:ha}).

Constraints~(\ref{eq:constraint1}) and~(\ref{eq:constraint2}) cannot be directly substituted into~(\ref{eq:ha}). However, we can use a similar approach as in the previous section for deriving~(\ref{eq:constraint1}) and~(\ref{eq:constraint2}). First, ideal $J$ generated by six polynomials~(\ref{eq:ha}),~(\ref{eq:constraint1}) and~(\ref{eq:constraint2}) is constructed. Then the unknown elements of the affine transformation $\Aff$ are eliminated from the generators of $J$.
We do this by computing the generators of 
$J_1 = J \cap \mathbb{C}[h_1,\dots,h_9,u_1,v_1,u_2,v_2,q_1,q_2,s_1,c_1,s_2,c_2]$.
The elimination ideal $J_1$ is generated by two polynomials:
\begin{eqnarray}
\label{eq:constraint_H_1}
h_8u_2s_1s_2+h_7u_2s_2c_1-h_8v_2s_1c_2-h_7v_2c_1c_2+\\
-h_2s_1s_2-h_1s_2c_1+h_5s_1c_2+h_4c_1c_2 = 0,
\nonumber
\\
\label{eq:constraint_H_2}
h_7^2u_1^2q_2+2h_7h_8u_1v_1q_2+h_8^2v_1^2q_2+h_5h_7u_2q_1+\\
\nonumber
-h_4h_8u_2q_1-h_2h_7v_2q_1+h_1h_8v_2q_1+2h_7h_9u_1q_2+\\
\nonumber
2h_8h_9v_1q_2+ h_2h_4q_1-h_1h_5q_1+h_9^2q_2 = 0.
\end{eqnarray}
The input code for {\tt Macaulay2} used to compute these generators is provided as supplementary material.

Polynomials~(\ref{eq:constraint_H_1}) and~(\ref{eq:constraint_H_2}) are new constraints that relate the homography matrix to the scales and rotations of the features in the first and second images. These constraints will help us for recovering $\Hom$ from two orientation- and scale-covariant feature correspondences.

\subsection{2-SIFT solver}

Constraint~(\ref{eq:constraint_H_1}) is linear in the elements of $\Hom$. For two SIFT correspondences, two such equations are given, which, together with the four equations for point correspondences~(\ref{eq:orig_dlt}), result in six homogeneous linear equations in the nine elements of $\Hom$.
In matrix form, these equations are: 
\begin{eqnarray}
\small
\matr{M}\,\matr{h} = \matr{0},
\end{eqnarray}
where $\matr{M}$ is a ${6}\times{9}$ coefficient matrix and $\matr{h}$ is a vector of 9 
elements of homography matrix $\Hom$.
For two  SIFT correspondences in two views, coefficient matrix $\matr{M}$ has a three-dimensional null space.
Therefore, the homography matrix can be parameterized by two unknowns as
\begin{eqnarray}
\small
\matr{H} = x\,\matr{H}_1 + y\,\matr{H}_2 + \matr{H}_3,
\label{eq:paramH}
\end{eqnarray}
where $\matr{H}_1,\matr{H}_2,\matr{H}_3$ are created from the 3D null space of $\matr{M}$ and $x$ and $y$ are new unknowns.
Now we can plug the parameterization~(\ref{eq:paramH}) into constraint~(\ref{eq:constraint_H_2}). For two SIFT correspondences, this 
results in two quadratic equations in two unknowns. Such equations have four solutions and they can be easily solved using e.g.\ the \gb or the resultant based method~\cite{cox2005}.
Here, we use the solver based on \gb method that can be created using the automatic generator~\cite{kukelovaECCV2008}. This solver performs Gauss-Jordan elimination of a ${6}\times{10}$ template matrix which contains just monomial multiples of the two input equations. Then the solver extracts solutions to $x$ and $y$ from the eigenvectors of a ${4}\times{4}$ multiplication matrix that is extracted from the template matrix. 
Finally, up to four real solutions to $\Hom$ are computed by substituting solutions for $x$ and $y$ to~(\ref{eq:paramH}).

Note that we do not know any degeneracies of the proposed solver which can occur in real life. For instance, the degeneracy of the four-point algorithm, i.e.\ the points are co-linear, is not a degenerate case for the 2SIFT solver.  

\subsection{Normalization of the affine parameters}

The normalization of the point coordinates is a crucial step to increase the numerical stability of $\Hom$ estimation~\cite{hartley1997defense}. 
Suppose that we are given a $3 \times 3$ normalizing transformation $\matr{T}_i$ transforming the center of gravity of the point cloud in the $i$th image to the origin and its average distance from it to $\sqrt{2}$. The formula for normalizing $\Aff$ is as follows~\cite{barath2018efficient}:
\vspace{-3.0mm} 

\begin{equation}
    \widehat{\Aff} = \matr{T}_2 
    \begin{bmatrix} 
        \Aff & 0 \\ 
        0 & 1
    \end{bmatrix}
    \matr{T}_1^{-1}, 
\end{equation}
where $\widehat{\Aff}$ is the normalized affinity. Matrix $\matr{T}_i$ transforms the points by translating them (last column) and applying a uniform scaling (diagonal).
Due to the fact that the last column of $\matr{T}_i$ has no effect on the top-left $2 \times 2$ sub-matrix of the normalized affinity, the equation can be rewritten as follows:
$
    \widehat{\Aff} = \text{diag}(t_2, t_2) \; \Aff \; \text{diag}(1 / t_1, 1 / t_1) = t_2 / t_1 \Aff,
$
where $t_1$ and $t_2$ are the scales of the normalizing transformations in the two images. 
Thus, for normalizing the affine transformation, it has to be multiplied by $t_2 / t_1$.

The scaling factor affects constraint~(\ref{eq:constraint1}) which, for $\widehat{\Aff}$, has the form
\begin{eqnarray}
\small
\label{eq:constraint1_scaled}
t^2q_1^2\widehat{a}_2\widehat{a}_3-t^2q_1^2\widehat{a}_1\widehat{a}_4+q_1q_2=0,
\end{eqnarray}
where $t = {t_1}/{t_2}$ and $\widehat{a_i}$ are elements of $\widehat{\Aff}$. Consequently constraint~(\ref{eq:constraint_H_2}) for the normalized coordinates has the form 
\begin{eqnarray}
\small
\label{eq:constraint_H_2_scaled}
h_7^2u_1^2q_2t^2+2h_7h_8u_1v_1q_2t^2+h_8^2v_1^2q_2t^2+h_5h_7u_2q_1+\\
\nonumber
-h_4h_8u_2q_1-h_2h_7v_2q_1+h_1h_8v_2q_1+2h_7h_9u_1q_2t^2+\\
\nonumber
2h_8h_9v_1q_2t^2+ h_2h_4q_1-h_1h_5q_1+h_9^2q_2t^2 = 0.
\end{eqnarray}
Note that this normalization does not affect the structure of the derived 2SIFT solver. The only difference is that, for the normalized coordinates, the coefficients in the template matrix are multiplied by scale factor $t$ as in~(\ref{eq:constraint_H_2_scaled}).

\begin{figure}[t] 
    \centering
	\includegraphics[width=0.55\columnwidth]{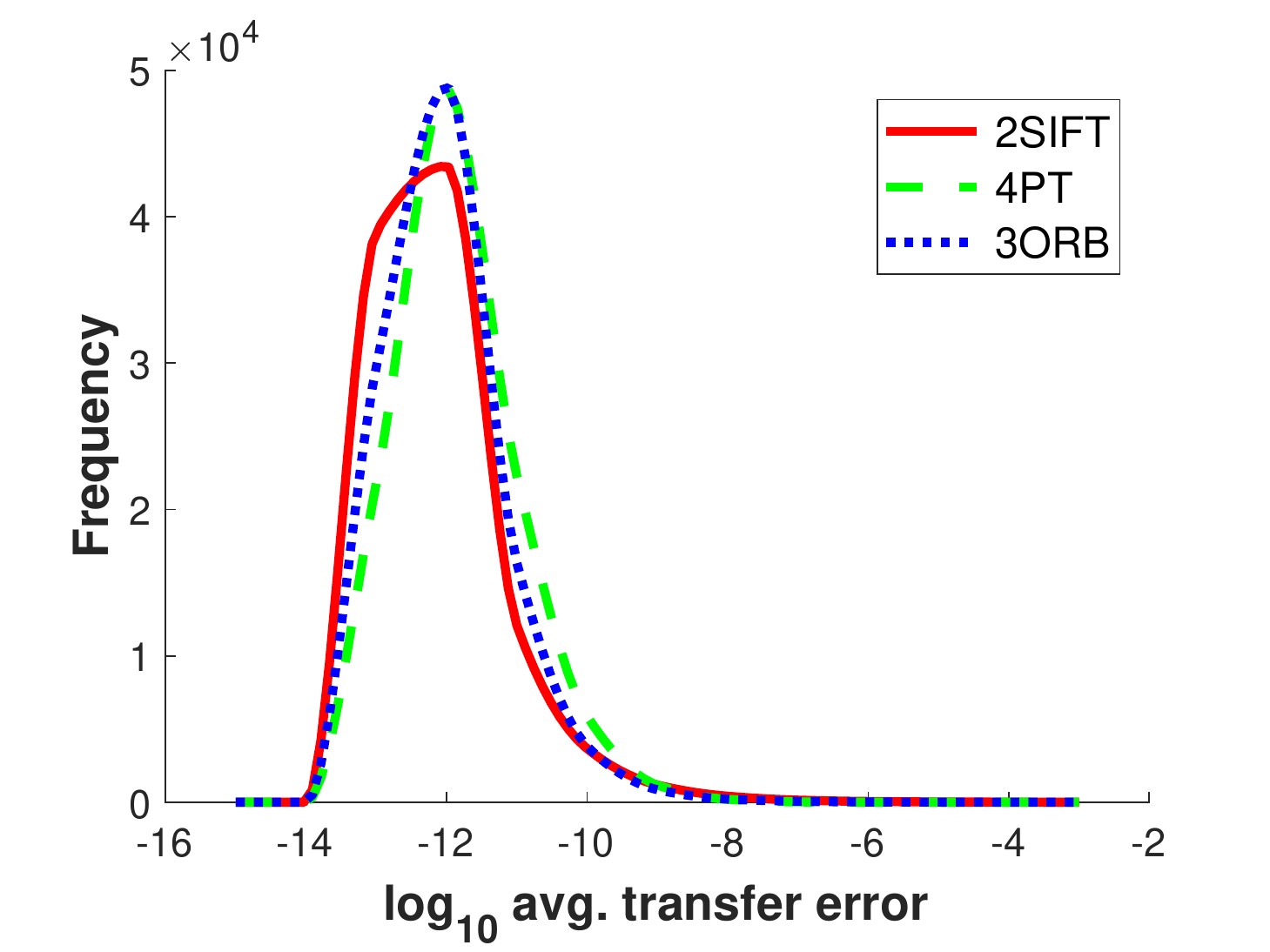}
    \caption{ \textit{Stability study.} The frequencies ($100\;000$ runs; vertical axis) of $\text{log}_{10}$ errors (horizontal) in the estimated homographies by the proposed (red), 4PT (green) and 3ORI (blue) methods. }
\label{fig:stability}
\end{figure} 

\section{Experimental results}

In this section, we compare the proposed solver (2SIFT) with the widely-used normalized four-point (4PT) algorithm~\cite{hartley2003multiple} and a method using three oriented features~\cite{barath2018five} (3ORI) for estimating the homography. 

\subsection{Computational complexity}

First, we compare the computational complexity of the competitor algorithms, see Table~\ref{tab:computational_complexity}. The first row consists of the major steps of each solver. For instance, $6 \times 9$ SVD + $6 \times 6$ QR + $4 \times 4$ EIG means, that the major steps are: the SVD decomposition of a $6 \times 9$ matrix, the QR decomposition of a $6 \times 6$ matrix and the eigendecomposition of a $4 \times 4$ matrix. 
In the second row, the implied computational complexities are summed. 
In the third one, the number of correspondences required for the solvers are written. 
The fourth row lists example outlier ratios in the data. 
In the fifth one, the theoretical number of iterations of RANSAC~\cite{hartley2003multiple} is written for each outlier ratio with confidence set to $0.99$.
The last row shows the computational complexity, i.e.\ the complexity of one iteration multiplied by the number of iteration, of RANSAC combined with the minimal methods. 
It can be seen that the proposed method leads to significantly smaller computational complexity. 
Moreover, we believe that by designing a specific solver to our two quadratic equations in two unknowns, similarly as in~\cite{kukelovaCVPR2016}, the computational complexity of our solver can be even reduced. 

\subsection{Synthesized tests}

\begin{table*}
	\center
	\resizebox{0.9\textwidth}{!}{\begin{tabular}{ c | c c c c | c c c c | c c c c }
    \hline
   		\cellcolor{black!10} & \multicolumn{4}{ c | }{ \cellcolor{black!10}\textbf{2SIFT} } & \multicolumn{4}{ c }{ \cellcolor{black!10}3ORI~\cite{barath2018five} } & \multicolumn{4}{ c }{ \cellcolor{black!10}4PT~\cite{hartley2003multiple} } \\
    \hline
   		steps & \multicolumn{4}{ c | }{$6 \times 9$ SVD + $6 \times 6$ QR + $4 \times 4$ EIG} & \multicolumn{4}{ c | }{$6 \times 9$ SVD } & \multicolumn{4}{ c }{$8 \times 9$ SVD}\\ 
   		1 iter & \multicolumn{4}{ c | }{$6 * 9^2 + 6^3 + 4^3 = 766$} & \multicolumn{4}{ c | }{$6 * 9^2 = 486$} & \multicolumn{4}{ c }{$8 * 9^2 = 649$}\\ 
    \hline
   		$m$ & \multicolumn{4}{ c | }{2} & \multicolumn{4}{ c | }{3} & \multicolumn{4}{ c }{4}\\
    \hline
   		1 - $\mu$ & 0.25 & 0.50 & 0.75 & 0.90 & 0.25 & 0.50 & 0.75 & 0.90 & 0.25 & 0.50 & 0.75 & 0.90 \\ 
    \hline 
       	\# iters & 6 & 16 & 71 & 458 & 8 & 34 & 292 & 4603 & 12 & 71 & 1177 & 46\;049 \\ 
       	\# comps & 4\;596 & 12\;256 & 54\;386 & 350\;828 & 3\;888 & 16\;524 & 141\;912 & 2\;237\;058 & 7\;788 & 46\;079 & 763\;873 & 29\;885\;801 \\ 
    \hline      
\end{tabular}}
\caption{The theoretical computational complexity of the solvers. The operations in the solvers ($1$st row -- steps), the computational complexity of one estimation ($2$nd -- 1 iter), the correspondence number required for the estimation ($3$rd -- $m$), possible outlier ratios ($4$th -- $1 - \mu$), the iteration number required for RANSAC with the confidence set to $0.95$ ($5$th -- \# iters), and computation complexity of the full procedure ($6$th -- \# comps).}
\label{tab:computational_complexity}
\end{table*}

\begin{figure*}[h]
	\includegraphics[height=0.510\columnwidth]{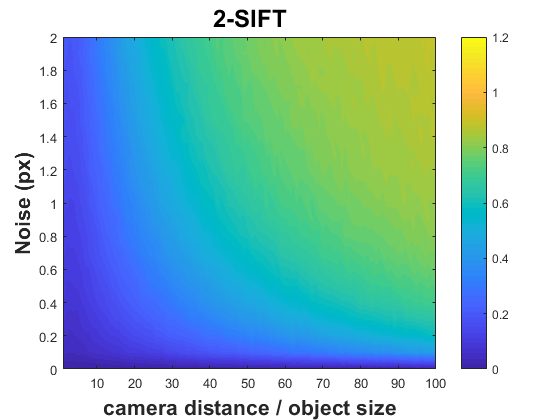}
	\includegraphics[height=0.510\columnwidth]{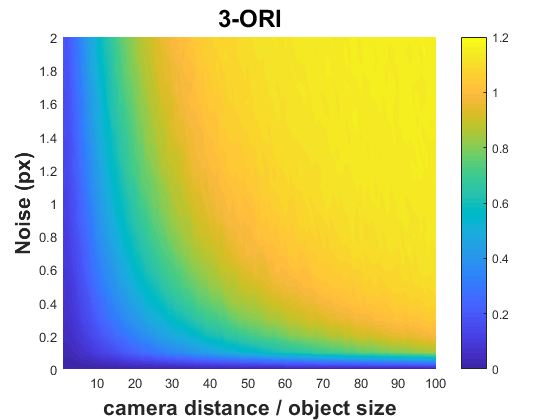}
	\includegraphics[height=0.510\columnwidth]{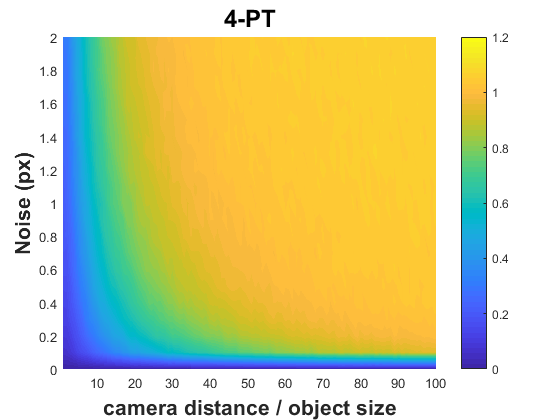}
\caption{ The average (of $10\;000$ runs on each noise $\sigma$) re-projection error of homography fitting to synthesized data by the proposed (2SIFT), normalized 4PT~\cite{hartley2003multiple} and 3ORI~\cite{barath2018five} methods. Each camera is located randomly on a center-aligned sphere. Ten points from the object are projected into the cameras, and zero-mean Gaussian-noise is added to the coordinates. The affine parameters are calculated from the noisy coordinates. 
The re-projection error (in px; shown by color) is plotted as the function of the "camera distance from the object / object size" ratio (horizontal) and the noise $\sigma$ (in px; vertical). }
\label{fig:synthetic_tests}
\end{figure*}

\begin{figure}[h]
    \centering
	\includegraphics[width=0.44\columnwidth]{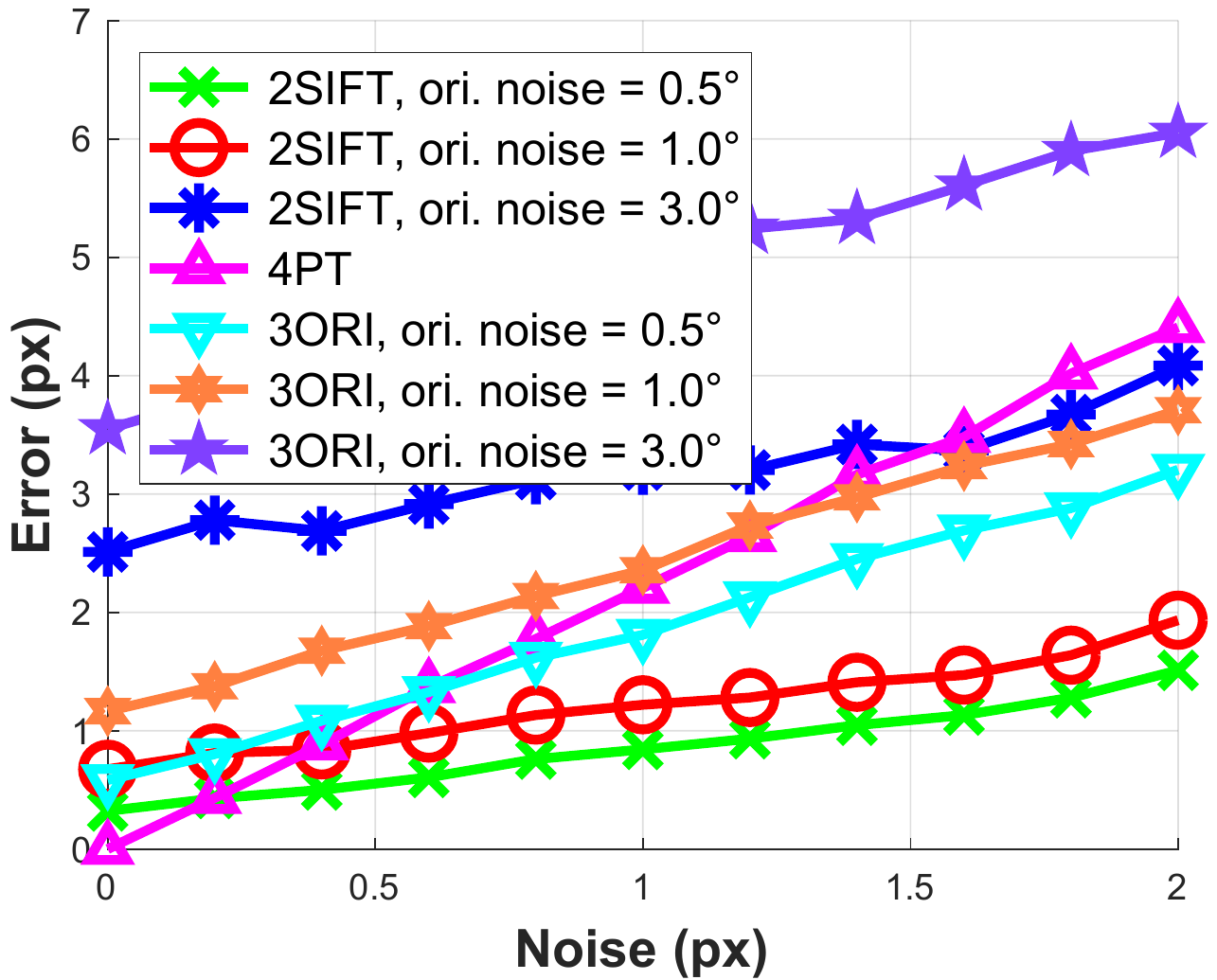}
	\includegraphics[width=0.44\columnwidth]{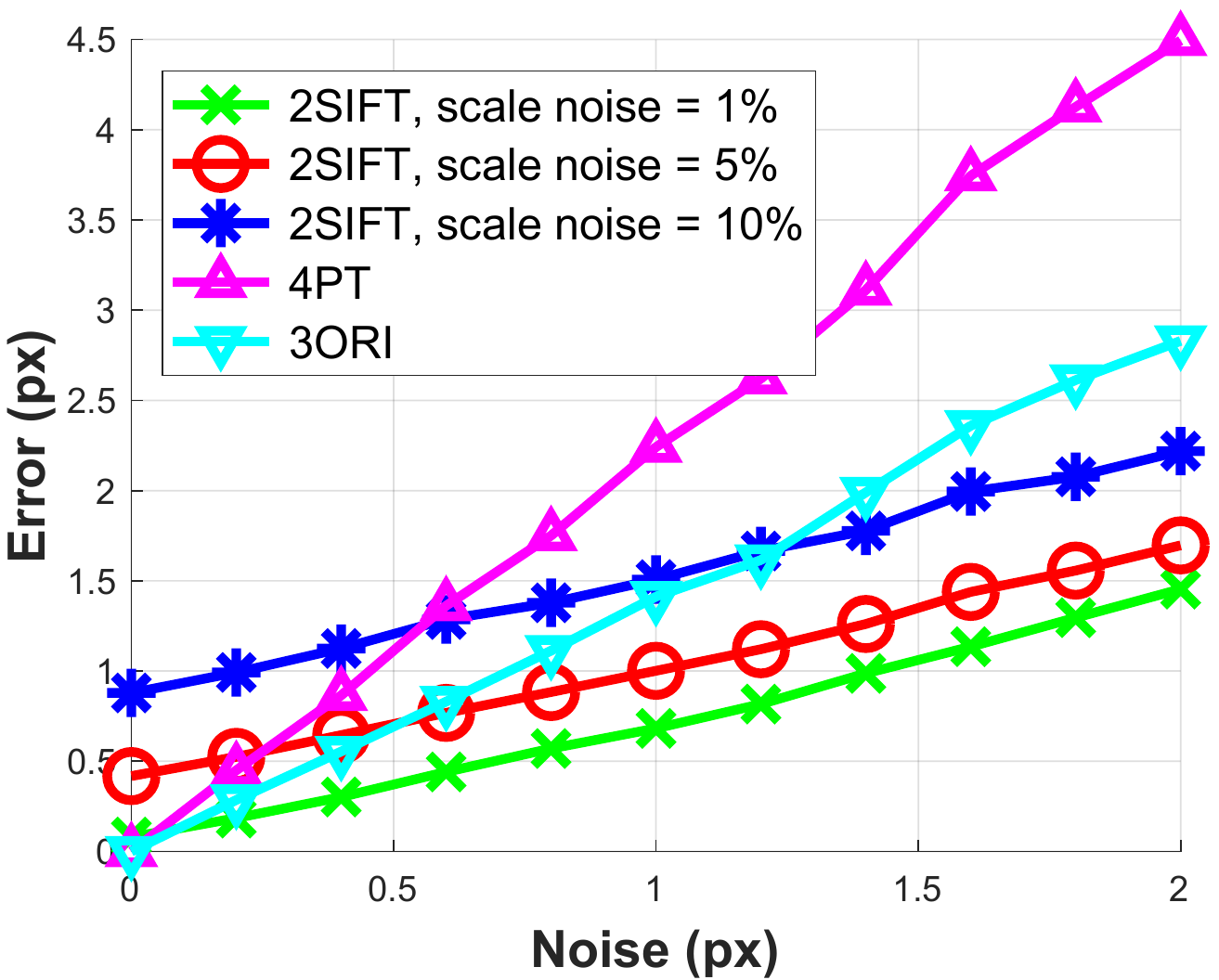}\\
	\includegraphics[width=0.44\columnwidth]{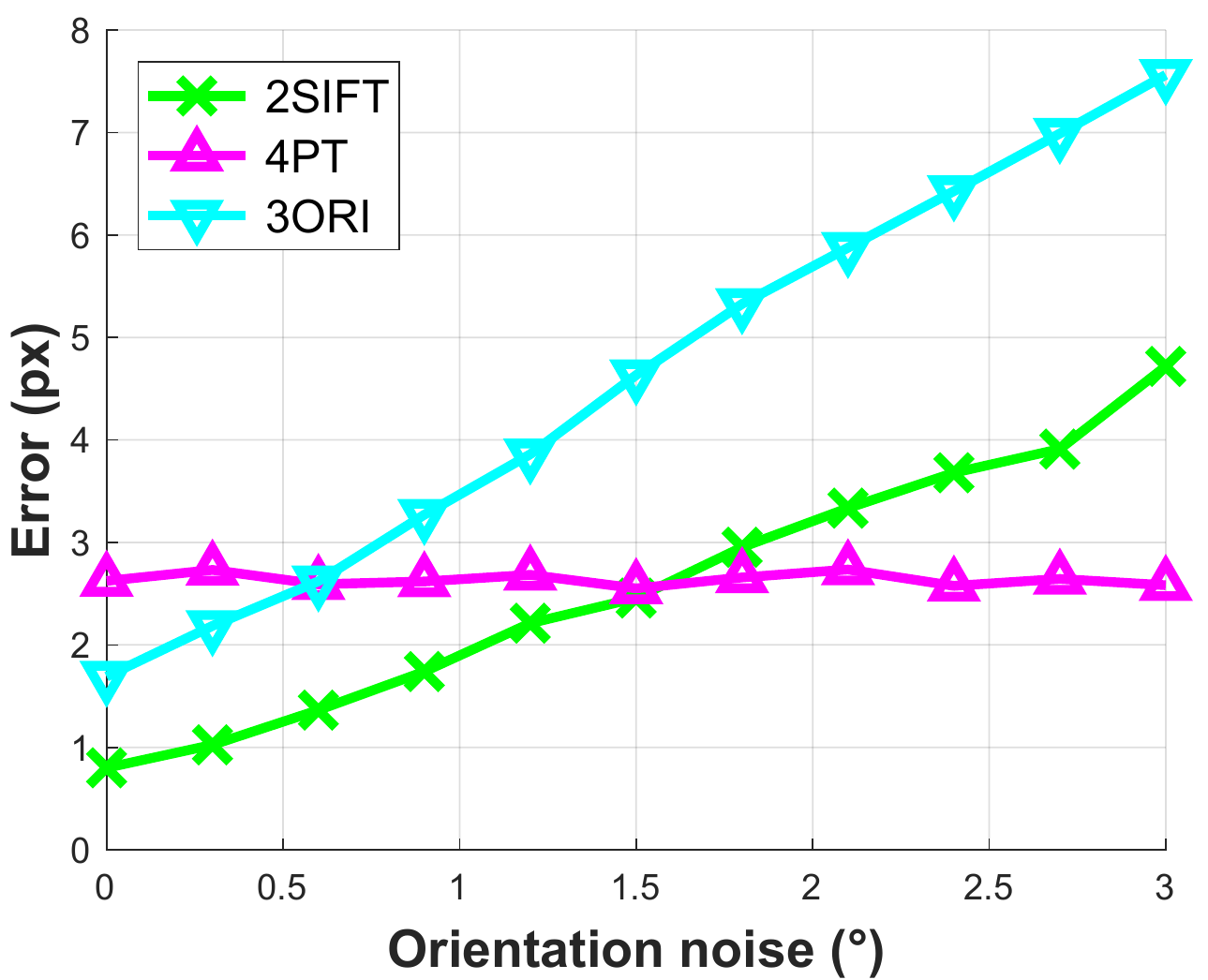}
	\includegraphics[width=0.44\columnwidth]{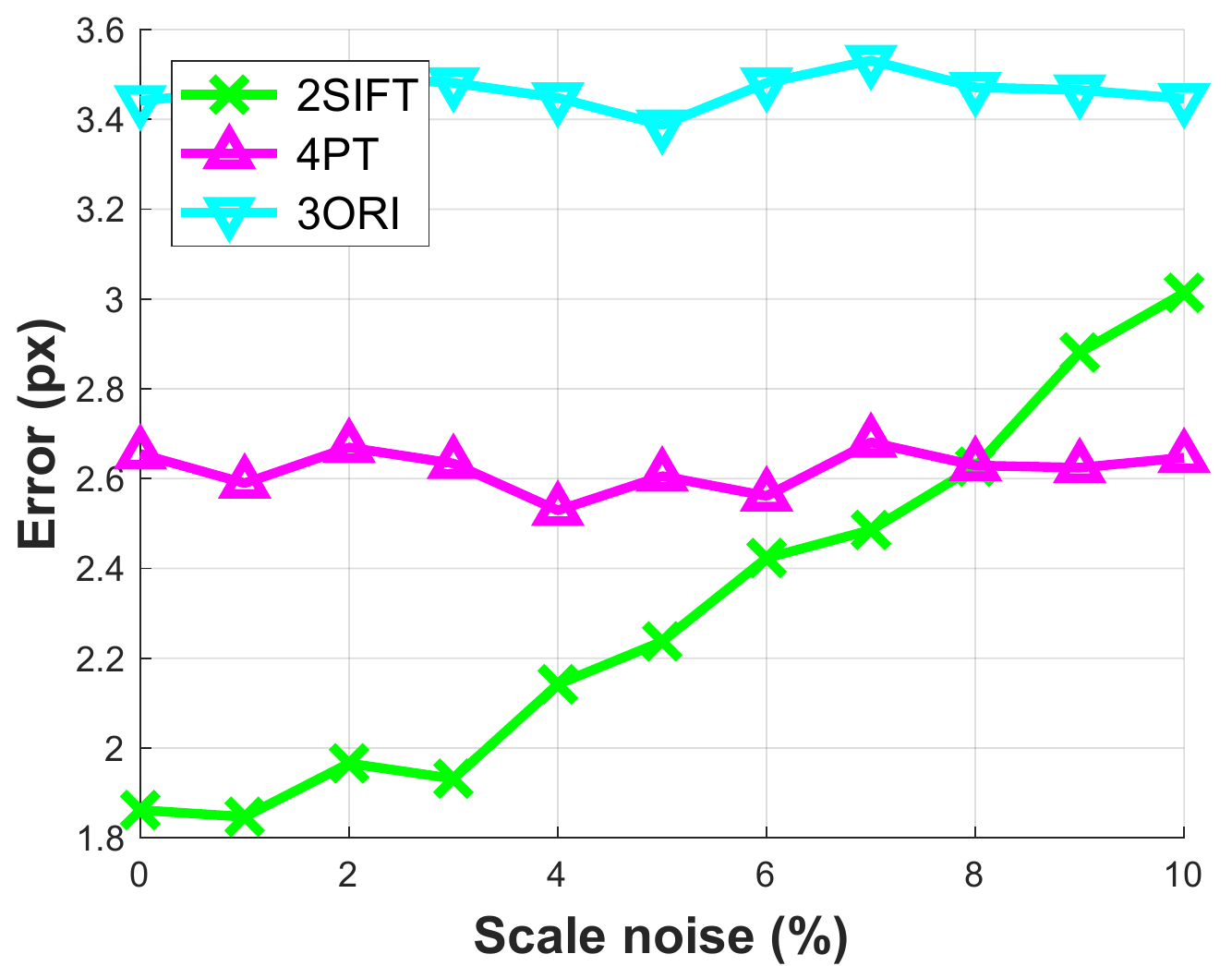}
\caption{ The average ($10\;000$ runs on each noise $\sigma$) re-projection error of homography fitting to synthesized data by the 2SIFT, normalized 4PT~\cite{hartley2003multiple} and 3ORI~\cite{barath2018five} methods. The same test scene is used as in Figure~\ref{fig:synthetic_tests}. For each plot, additional noise was added to the orientations or the scales besides the noise coming from the noisy affine transformations. 
(\textit{Top}) The error is plotted as the function of the image noise $\sigma$. The curves show the results on different noise levels in the orientations and scales.
(\textit{Bottom}) The error is plotted as the function of the orientation (left plot) and scale (right) noise. The noise in the point coordinates was set to $1.0$ px. The scale noise for the left plot was set to $1\%$. The orientation noise for the right one was set to \ang{1}. }
\label{fig:synthetic_tests_corrupted_sift}
\end{figure}

\begin{figure*}[h]
    \centering
	\includegraphics[width=0.60\columnwidth]{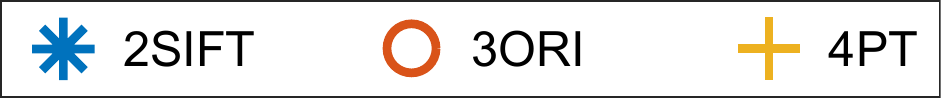}\\
	\begin{subfigure}[t]{2.0\columnwidth}
	    \includegraphics[width=0.330\columnwidth]{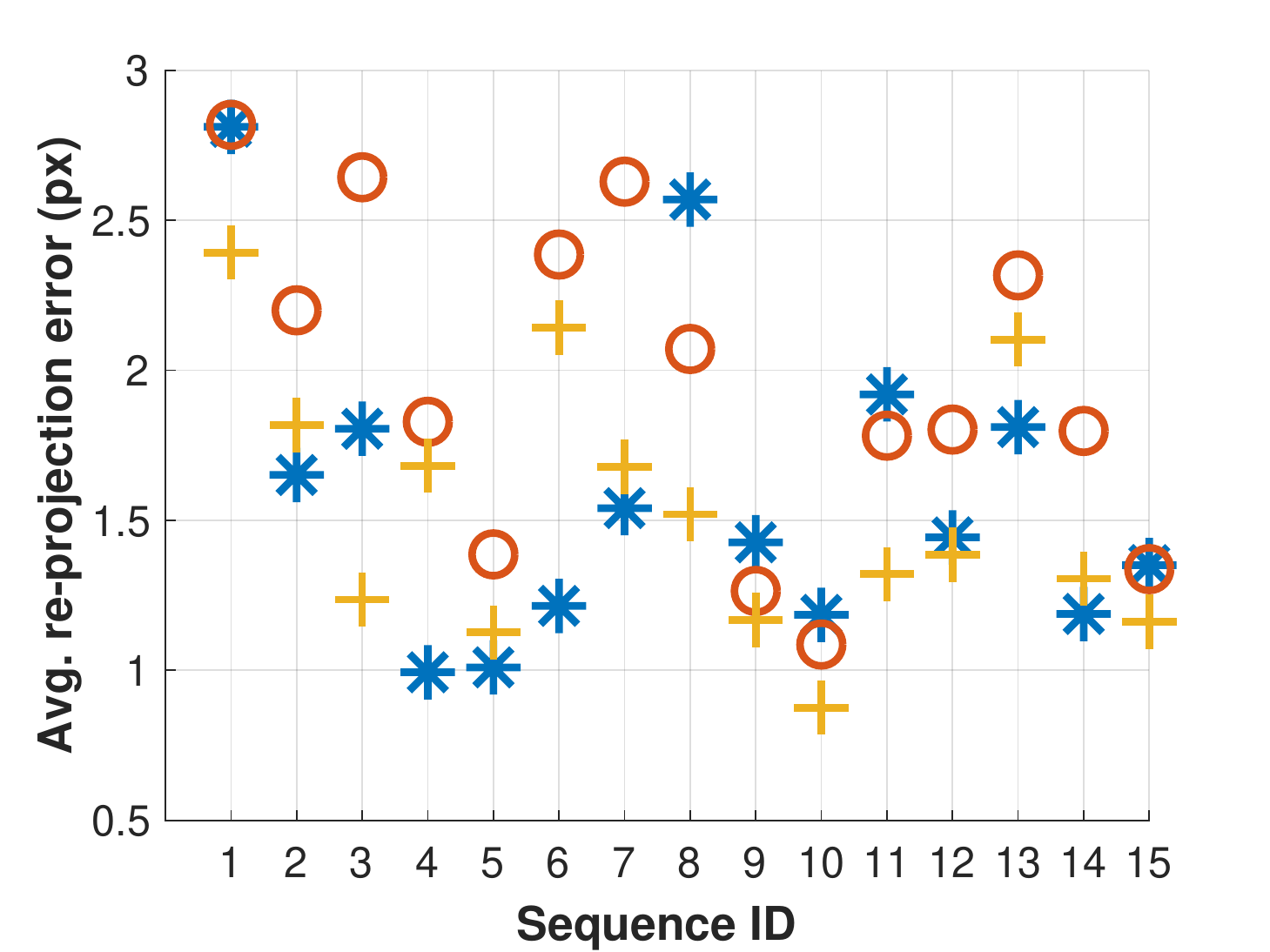}
	    \includegraphics[width=0.330\columnwidth]{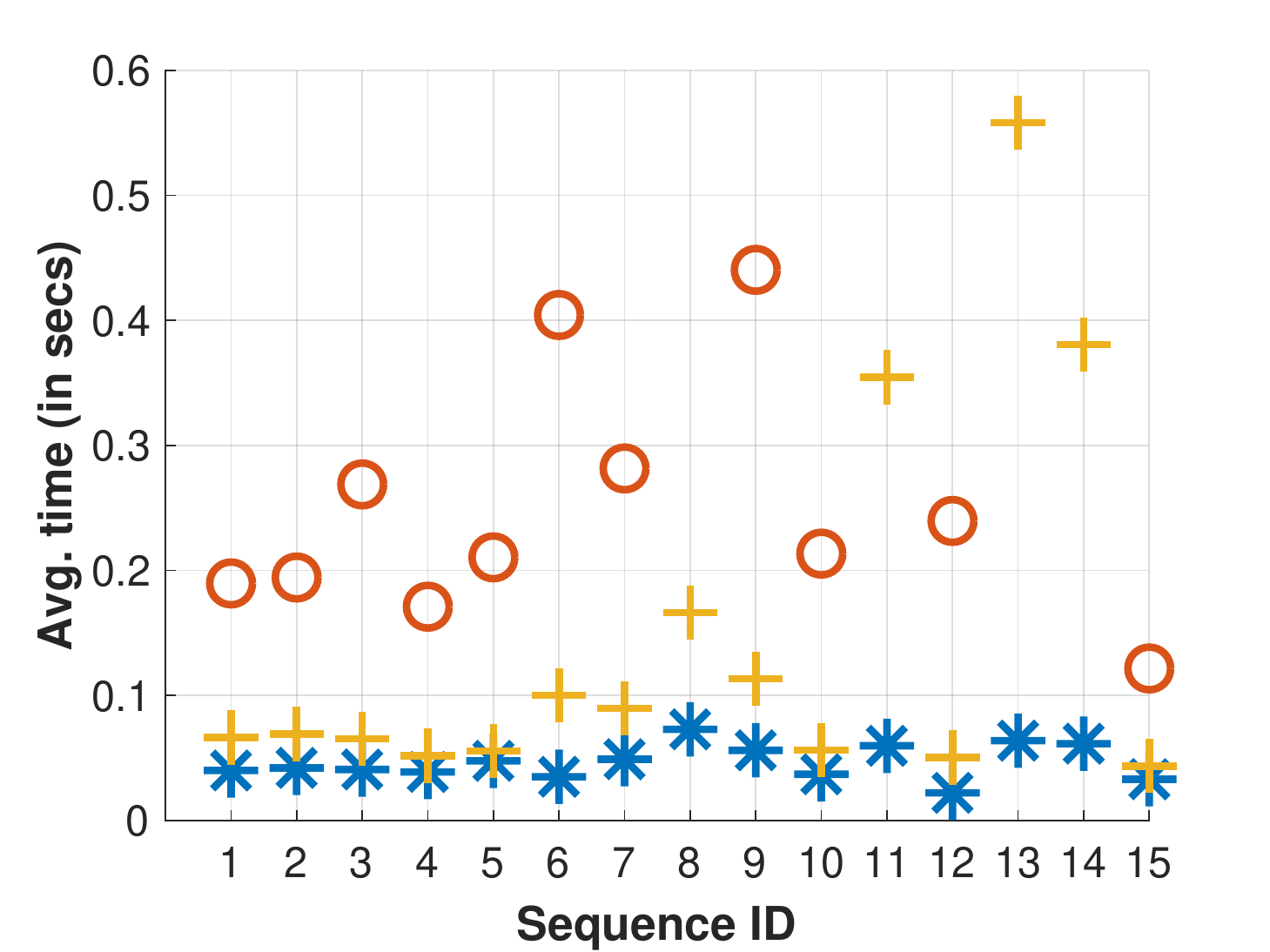}
	    \includegraphics[width=0.330\columnwidth]{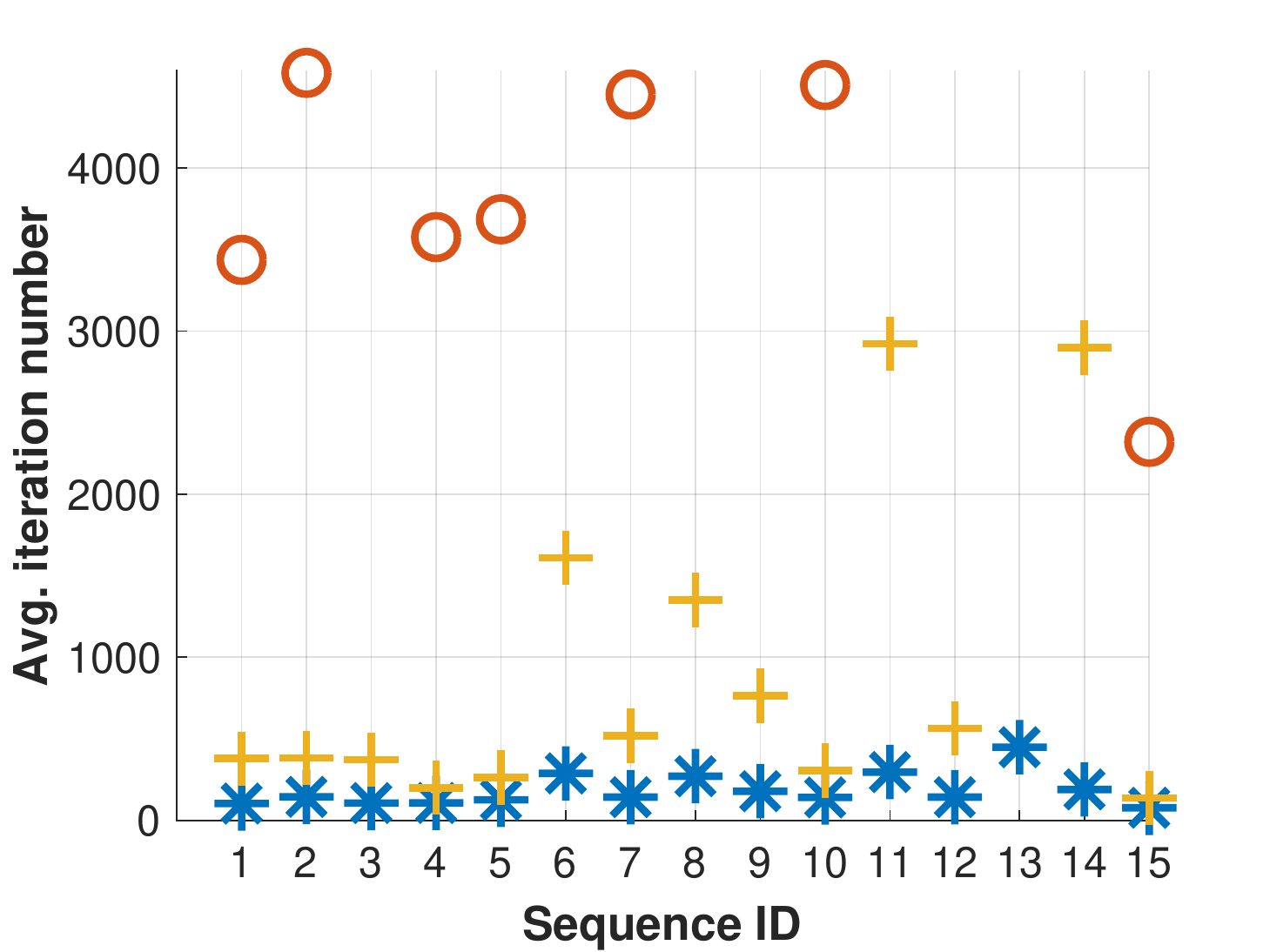}
        \label{fig:malaga_results_a}
	\end{subfigure}
\caption{ The results on $15$ sequences ($9\;064$ image pairs) of the {\fontfamily{cmtt}\selectfont Malaga} dataset using GC-RANSAC~\cite{barath2017graph} as a robust estimator and different minimal solvers (2SIFT, 3ORI, 4PT). The confidence of RANSAC was set to $0.95$ and the inlier-outlier threshold to $2.0$ pixels. The re-projection error (left; in pixels), average processing time (middle; in seconds) and average iteration number (right) are reported. }
\label{fig:malaga_results}
\end{figure*}

To test the accuracy of the homographies obtained by the proposed method, first, we created a synthetic scene consisting of two cameras represented by their $3 \times 4$ projection matrices $\matr{P}_1$ and $\matr{P}_2$. 
They were located randomly on a center-aligned sphere. 
A plane with random normal was generated in the origin and ten random points, lying on the plane, were projected into both cameras. The points were at most one unit far from the origin.
To get the ground truth affine transformations, we calculated homography $\Hom$ by projecting four random points from the plane to the cameras and applying the normalized DLT~\cite{hartley2003multiple} algorithm. 
The local affine transformation of each correspondence was computed from the ground truth homography by (\ref{eq:taylor_approximation}). 
Note that $\Hom$ could have been calculated directly from the plane parameters. 
However, using four points promised an indirect but geometrically interpretable way of noising the affine parameters: adding noise to the coordinates of the four projected points. 
To simulate the SIFT orientations and scales, $\Aff$ was decomposed to $\matr{J}_1$, $\matr{J}_2$. 
Since the decomposition is ambiguous, $\alpha_1$, $q_{u,1}$, $q_{v,1}$, $w_1$ were set to random values. $\matr{J}_1$ was calculated from them. Finally, $\matr{J}_2$ was calculated as $\matr{J}_2 = \Aff \matr{J}_1$.
Zero-mean Gaussian-noise was added to the point coordinates, and, also, to the coordinates which were used to estimate the affine transformations.

Fig.~\ref{fig:stability} reports the numerical stability of the methods in the noise-free case. The frequencies (vertical axis), i.e.\ the number of occurrences in $100\;000$ runs, are plotted as the function of the $\log_{10}$ average transfer error (in px; horizontal) computed from the estimated homography and the not used correspondences. 
It can be seen that all tested solvers are numerically stable.
Fig.~\ref{fig:synthetic_tests} plots the $|| \textbf{H}_{\text{est}} - \textbf{H}_{\text{gt}} ||_\text{F}$ errors as the function of image noise level $\sigma$ (vertical axis) and the ratio (horizontal) of the camera distance, i.e.\ the radius of the sphere on which the cameras lie, and the object size. The homographies were normalized. The proposed 2SIFT algorithm (left) is less sensitive to the choice of both parameters than the 3ORI (middle) and 4PT (right) methods. 

Fig.~\ref{fig:synthetic_tests_corrupted_sift} reports the re-projection error (vertical; in pixels) as the function of the image noise $\sigma$ with additional noise added to the SIFT orientations (left) and scales (right) besides the noise coming from the noisy affine transformations. 
In the top row, the error is plotted as the function of the image noise $\sigma$. 
The curves show the results on different noise levels in the orientations and scales.
In the bottom row, the error is plotted as the function of the orientation (left plot) and scale (right) noise. 
The noise in the point coordinates was set to $1.0$ px. The scale noise for the left plot was set to $1\%$. The orientation noise for the right one was set to \ang{1}.
It can be seen that, even for large noise in the scale and orientation, the new solver performs reasonably well.

\subsection{Real world tests}

To test the proposed method on real-world data, we downloaded the 
{\fontfamily{cmtt}\selectfont AdelaideRMF}\footnote{{cs.adelaide.edu.au/~hwong/doku.php?id=data}}, 
{\fontfamily{cmtt}\selectfont Multi-H}\footnote{{web.eee.sztaki.hu/~dbarath}}, {\fontfamily{cmtt}\selectfont Malaga}\footnote{{www.mrpt.org/MalagaUrbanDataset}} and {\fontfamily{cmtt}\selectfont Strecha}\footnote{{https://cvlab.epfl.ch/}} 
datasets. 
{\fontfamily{cmtt}\selectfont AdelaideRMF} and {\fontfamily{cmtt}\selectfont Multi-H} consist of image pairs of resolution from $455 \times 341$ to $2592 \times 1944$ and manually annotated (assigned to a homography or to the outlier class) correspondences. Since the reference point sets do not contain rotations and scales, we detected points applying the SIFT detector. 
The correspondences provided in the datasets were used to estimate ground truth homographies.
For each homography, we selected the points out of the detected SIFT correspondences which are closer than a manually set inlier-outlier threshold, i.e.\ $2$ pixels. 
As robust estimator, we chose GC-RANSAC~\cite{barath2017graph} since it is state-of-the-art and its implementation is available\footnote{\url{https://github.com/danini/graph-cut-ransac}}. 
GC-RANSAC is a locally optimized RANSAC with PROSAC~\cite{chum2005matching} sampling.
For fitting to a minimal sample, GC-RANSAC used one of the compared methods, e.g.\ the proposed one. 
For fitting to a non-minimal sample, the normalized 4PT algorithm was applied. 

Given an image pair, the procedure to evaluate the estimators on {\fontfamily{cmtt}\selectfont AdelaideRMF} and {\fontfamily{cmtt}\selectfont Multi-H} is as follows: 
first, the ground truth homographies, estimated from the manually annotated correspondence sets, were selected one by one. For each homography:
(i) The correspondences which did not belong to the selected homography were replaced by completely random correspondences to reduce the probability of finding a different plane than what was currently tested. 
(ii) GC-RANSAC was applied to the point set consisting of the inliers of the homography and outliers. 
(iii) The estimated homography is compared to the ground truth one estimated from the manually selected inliers.
 
\begin{table}[t]
\center
  	\resizebox{1.0\columnwidth}{!}{\begin{tabular}{| c | r | r r r | }
    \hline    
 	 	 & & \cellcolor{Gray} 2SIFT & 3ORI~\cite{barath2018five} & 4PT~\cite{hartley2003multiple} \\ 
    \hline
    	  & $\epsilon$ (px) & \cellcolor{Gray} \textbf{1.57} & 1.97 & 1.61 \\ 
    	 \small AdelaideRMF & \ph{x}\# iters. & \cellcolor{Gray} \textbf{877} & 9\;772 & 26\;082 \\ 
    	 \small ($43$\#) & time (s) & \cellcolor{Gray} \textbf{0.092} & 0.918 & 2.989 \\
    \hline
    	  & $\epsilon$ (px) & \cellcolor{Gray} 1.90 & 3.41 & \textbf{1.87} \\ 
    	 \small Multi-H & \ph{x}\# iters. & \cellcolor{Gray} \textbf{80\;031} & 458\;800 & 410\;781 \\ 
    	 \small ($33$\#) & time (s) & \cellcolor{Gray} \textbf{57.921} & 213.900 & 300.645 \\
    \hline
    	  & $\epsilon$ (px) & \cellcolor{Gray} 1.42 & 1.51 & \textbf{1.25} \\ 
    	 \small Strecha & \ph{x}\# iters. & \cellcolor{Gray} \textbf{4\;718} & 17\;414 & 60\;973 \\ 
    	 \small ($852$\#) & time (s) & \cellcolor{Gray} \textbf{1.435} & 3.180 & 10.246 \\
    \hline  
\end{tabular}}
\caption{ Homography estimation on the {\fontfamily{cmtt}\selectfont AdelaideRMF} ($18$ pairs; $43$ planes) and {\fontfamily{cmtt}\selectfont Multi-H} ($4$ pairs; $33$ planes) and {\fontfamily{cmtt}\selectfont Strecha} datasets ($852$ planes) by GC-RANSAC~\cite{barath2017graph} combined with minimal methods. Each column reports the results of a method. The required confidence was set to $0.95$. The reported properties are the mean re-projection error ($\epsilon$, in pixels); the number of samples drawn by GC-RANSAC (\# iters.); and the processing time in seconds. Average of $100$ runs on each image pair.}
\label{tab:dataset_comparison}
\end{table}

The {\fontfamily{cmtt}\selectfont Strecha} dataset consists of image sequences of buildings. 
All images are of size $3072 \times 2048$. 
The methods were applied to all possible image pairs in each sequence.
The {\fontfamily{cmtt}\selectfont Malaga} dataset was gathered entirely in urban scenarios with a car equipped with several sensors, including a high-resolution camera and five laser scanners. $15$ video sequences are provided and we used every $10$th image from each sequence. 
The ground truth projection matrices are provided for both datasets.
To get a reference correspondence set for each image pair in the {\fontfamily{cmtt}\selectfont Strecha} and {\fontfamily{cmtt}\selectfont Malaga} datasets, first, calculated the fundamental matrix from the ground truth camera poses provided in the datasets. 
SIFT detector was applied.
Correspondences were selected for which the symmetric epipolar distance was smaller than $1.0$ pixel. 
RANSAC was applied to the filtered correspondences finding the most dominant homography with a threshold set to $1.0$ pixel and confidence to $0.9999$. 
The inliers of this homography were considered as a reference set. 
In case of having less then $50$ reference points, the pair was discarded from the evaluation. 
In total, $852$ image pairs were tested in the {\fontfamily{cmtt}\selectfont Strecha} dataset and $9\;064$ pairs in the {\fontfamily{cmtt}\selectfont Malaga} dataset. 

Example results are shown in Fig.~\ref{fig:example_image_pairs}. The inliers of the homography estimated by estimated by 2SIFT are drawn. Also, the number of iteration required for 2SIFT and 4PT and the ground truth inlier ratios are reported. In all cases, \textit{2SIFT made significantly fewer iterations} than 4PT.  

Table~\ref{tab:dataset_comparison} reports the results on the {\fontfamily{cmtt}\selectfont AdelaideRMF} (rows 2--4), {\fontfamily{cmtt}\selectfont Multi-H} (5--7) and {\fontfamily{cmtt}\selectfont Strecha} (8--10) datasets. 
The names of the datasets are written into the first column and the numbers of planes are in brackets. 
The names of the tested techniques are written in the first row. 
Each block, consisting of three rows, shows the mean re-projection error computed from the manually annotated correspondences and the estimated homographies ($\epsilon$; in pixels; avg. of $100$ runs on each pair); the number of samples drawn by the outer loop of GC-RANSAC ($\#$ iters.); and the processing time (in secs). 
The RANSAC confidence was set to $0.95$ and the inlier-outlier threshold to $2$ pixels. 
It can be seen that the proposed method has similar errors to that of the 4PT algorithm, but \textit{2SIFT leads to 1--2 orders of magnitude speedup} compared to 4PT. 

The results on the {\fontfamily{cmtt}\selectfont Malaga} dataset are shown in Figure~\ref{fig:malaga_results}.
The confidence of GC-RANSAC was set to $0.95$ and the inlier-outlier threshold to $2.0$ pixels. 
The reported properties are the average re-projection error (left; in pixels), processing time (middle; in seconds) and the average number of iterations (right). 
It can be seen that the re-projection errors of 4PT and 2SIFT are fairly similar
However, \textit{2SIFT is significantly faster in all cases} due to making much fewer iterations than 4PT. 

\section{Conclusion}

We proposed a theoretically justifiable interpretation of the angles and scales which the orientation- and scale-covariant feature detectors, e.g.\ SIFT or SURF, provide.
Building on this, two new general constraints are proposed for covariant features. 
These constraints are then exploited to derive two new formulas for homography estimation. 
Using the derived equations, a solver is proposed for estimating the homography from two correspondences. 
The new solver is numerically stable and easy to implement.
Moreover, it leads to results superior in terms of geometric accuracy in many cases. 
Also, it is shown how the normalization of the point correspondences 
affects the rotation and scale parameters.
Due to requiring merely two feature pairs, robust estimators, e.g.\ RANSAC, do significantly fewer iterations than by using the four-point algorithm. The method is tested in a synthetic environment and on publicly available real-world datasets consisting of thousands of image pairs. 
The source code is uploaded as supplementary material.

\appendix
\section{Proof the affine decomposition}
\label{app:proof_jacobians}

We prove that decomposition $\Aff = \matr{J}_2 \matr{J}_1^{-1}$, where $\matr{J}_i$ is the Jacobian of the projection function w.r.t.\ the directions in the $i$th image, is geometrically valid. 
Suppose that a three-dimensional point $\matr{P} = \begin{bmatrix} x & y & z \end{bmatrix}^\trans$ lying on a continuous surface $S$ is given. Its projection in the $i$th image is $\textbf{p}_i = \begin{bmatrix} u_i & v_i \end{bmatrix}^\trans$. The projected coordinates, $u_i$ and $v_i$, are determined by the projection functions $\matr{\Pi}_u, \matr{\Pi}_v: \mathbb{R}^3 \to \mathbb{R}$ as follows:  
$u_i = \mathbf{\Pi}^i_u(x,y,z)$, $v_i = \mathbf{\Pi}^i_v(x,y,z),$
where the coordinates of the surface point are written in parametric form as
%
	$x = \mathcal{X}(u, v)$, $y = \mathcal{Y}(u, v)$, $z = \mathcal{Z}(u, v)$.
%
%
It is well-known in differential geometry~\cite{kreyszig1968introduction} that the basis of the tangent plane at point $\textbf{P}$ is written by the partial derivatives of $S$ w.r.t.\ the spatial coordinates. The surface normal $\textbf{n}$ is expressed by the cross product of the tangent vectors $\textbf{s}_u$ and $\textbf{s}_v$ where
%
$
	\textbf{s}_u = \begin{bmatrix} 
    	\frac{\partial \mathcal{X}(u,v)}{\partial u} & 
    	\frac{\partial \mathcal{Y}(u,v)}{\partial u} &
    	\frac{\partial \mathcal{Z}(u,v)}{\partial u}
    \end{bmatrix}^\trans,
$
%
and $\textbf{s}_v$ is calculated similarly.
Finally, $\textbf{n} = \textbf{s}_u \times \textbf{s}_v$. 
Locally, around point $\textbf{P}$, the surface can be approximated by the tangent plane, therefore, the neighboring points in the $i$th image are written as the first-order Taylor-series as follows:
\begin{eqnarray*}
    \small
	\textbf{p}_i + \mathbf{\Delta} 
	\begin{bmatrix}
		\Pi_x(x, y, z) \\ \Pi_y(x, y, z)
	\end{bmatrix} + 
	\begin{bmatrix}
		\frac{\partial \Pi_x^i(x, y, z)}{\partial u} & \frac{\partial \Pi_x^i(x, y, z)}{\partial v} \\ 
       \frac{\partial \Pi_y^i(x, y, z)}{\partial u} & \frac{\partial \Pi_y^i(x, y, z)}{\partial v}
	\end{bmatrix} 
	\begin{bmatrix}
    	\Delta u \\
    	\Delta v
	\end{bmatrix}, 
\end{eqnarray*}
where $[\Delta v, \Delta u]^\trans$ is the translation on surface $S$, and $\Delta x$, $\Delta y$ are the coordinates of the implied translation added to $\textbf{p}_i$. 
It can be seen that transformation $\mathbf{J}_i$ mapping the infinitely close vicinity around point $\textbf{p}_i$ in the $i$th image is given as
\begin{equation*}
    \small
	\mathbf{J}_i = \begin{bmatrix}
		\frac{\partial \Pi_x^i(x, y, z)}{\partial u} & \frac{\partial \Pi_x^i(x, y, z)}{\partial v} \\ 
       \frac{\partial \Pi_y^i(x, y, z)}{\partial u} & \frac{\partial \Pi_y^i(x, y, z)}{\partial v}
	\end{bmatrix},
\end{equation*}
thus
\begin{equation*}
    \small
	\begin{bmatrix}
		\Delta x & \Delta y
	\end{bmatrix}^\trans \approx \textbf{J}_i \begin{bmatrix}
		\Delta u & \Delta v
	\end{bmatrix}^\trans.
\end{equation*}
The partial derivatives are reformulated using the chain rule. As an example, the first element it is as 
\begin{eqnarray*}
    \small
	\frac{\partial \Pi_x^i(x, y, z)}{\partial u} = 
    \frac{\partial \Pi_x^i(x, y, z)}{\partial x} \frac{x}{\partial u} + \\
    \frac{\partial \Pi_x^i(x, y, z)}{\partial x} \frac{y}{\partial u} + 
    \frac{\partial \Pi_x^i(x, y, z)}{\partial x} \frac{z}{\partial u} = \nabla (\mathbf{\Pi}_x^i)^\trans \mathbf{s}_u,
\end{eqnarray*}
where $\mathbf{\nabla \Pi}_x^i$ is the gradient vector of $\mathbf{\Pi}_x$ w.r.t.\ coordinates $x$, $y$ and $z$. Similarly,
\begin{eqnarray*}
    \small
	\frac{\partial {\Pi}_x^i}{\partial v} = \nabla (\mathbf{\Pi}_x^i)^\trans \textbf{s}_v, \;
    \frac{\partial {\Pi}_y^i}{\partial u} = \nabla (\mathbf{\Pi}_y^i)^\trans \textbf{s}_u, \;
    \frac{\partial {\Pi}_y^i}{\partial v} = \nabla (\mathbf{\Pi}_y^i)^\trans \textbf{s}_v, 
\end{eqnarray*}
Therefore, $\mathbf{J}_i$ can be written as
\begin{equation*}
    \small
	\mathbf{J}_i = \begin{bmatrix}
		\nabla (\mathbf{\Pi}_x^i)^\trans \\ 
		\nabla (\mathbf{\Pi}_y^i)^\trans
	\end{bmatrix}
    \begin{bmatrix}
		\textbf{s}_u & \textbf{s}_v
	\end{bmatrix}.
\end{equation*}
Local affine transformation $\mathbf{A}$ transforming the infinitely close vicinity of point $\textbf{p}_1$ in the first image to that of $\textbf{p}_2$ in the second one is as follows:
\begin{equation*}
    \small
	\begin{bmatrix}
		\Delta x_2 \\
		\Delta y_2
	\end{bmatrix} = 
	\mathbf{J}_2 \mathbf{J}_1^{-1} \begin{bmatrix}
		\Delta x_1 \\
		\Delta y_1
	\end{bmatrix} = 
	\mathbf{A} \begin{bmatrix}
		\Delta x_1 \\
		\Delta y_1
	\end{bmatrix}.
\end{equation*}

{\small
\bibliographystyle{ieee}
\bibliography{egbib}
}

\end{document}